\documentclass[11pt]{article}

\usepackage[margin=1in]{geometry}
\usepackage{amsmath, amssymb}
\usepackage{booktabs}
\usepackage{graphicx}
\usepackage{multirow}
\usepackage[round]{natbib}
\usepackage{xcolor}
\usepackage{hyperref}
\usepackage{titlesec}
\usepackage{authblk}
\usepackage[expansion=false]{microtype}

\titleformat{\section}{\normalfont\Large\bfseries}{\thesection}{1em}{}
\titleformat{\subsection}{\normalfont\large\bfseries}{\thesubsection}{1em}{}

\hypersetup{colorlinks=true, linkcolor=black, citecolor=blue!60!black, urlcolor=blue!60!black}

\usepackage[htt]{hyphenat}

\title{\textbf{Constraint-Aware Test-Time Scoring for\\LLM and VLM Candidate Selection}}

\author[1]{Rangarajan Ramesh\thanks{\texttt{rangarajan.ramesh@gmail.com}}}
\author[2]{Yongchao Huang\thanks{\texttt{yongchao.huang@abdn.ac.uk}}}
\affil[1]{Mitos Robotics}
\affil[2]{School of Natural and Computing Sciences, University of Aberdeen}
\date{}

\begin{document}
	\maketitle
	
	\begin{abstract}
		\noindent When an LLM or VLM produces several candidate answers, picking the best one usually relies on the model's own confidence, a learned judge model, or a domain check. Each has a known weakness: an answer can be fluent but insecure, mathematically plausible but physically infeasible, or rated highly by a judge while quietly breaking a task requirement. We propose a scoring framework that combines a small set of hard checks every answer must pass with a wider set of softer checks whose scores are blended into one number per candidate. This adaptive scoring framework consists of three components: a drafting model that produces the candidates, a configuring model that selects and weights the checks for each task, and scoring models that rate properties no grounded tool can compute. We apply our proposed method to four tasks: writing code that is both correct and secure (CWEval), generating robot manipulation plans (LIBERO with added safety variants), passing competitive-programming hidden tests (LiveCodeBench), and answering questions from retrieved documents (HotpotQA). An ablation study isolates the contribution of each component and finds that the largest gains come from combining hard checks with candidate comparison: on tasks where the first sampled answer is often broken or unsafe, this pairing lifts joint correctness-and-safety by up to $+0.21$ (retrieval-augmented QA) and code-generation safety by up to $+0.28$ (CWEval), with the size of the gain tracking how weak the first sample is, falling to near zero or slightly negative where it is already strong. The per-task layer that selects and weights the softer checks adds no measurable value over simpler alternatives. This builds a unified scoring framework, together with an ablation that identifies the grounded hard checks, not the soft-factor configuration, as the source of the gain.
	\end{abstract}
	
	\section{Introduction}
	
	LLMs and VLMs are increasingly deployed in scenarios where wrong outputs carry real consequences: code that runs against production systems, plans that drive physical hardware, factual claims that inform decisions. The failure modes that matter are not surface errors but confident outputs that look right and act wrong.
	
	To solve this, there are commonly two recipes. The first is to ground the output against something other than the model: compilers and unit tests for code, symbolic planners and collision checks for robot plans, retrieval and entailment for factual claims. These are reliable where they apply, but each is narrow. A compiler verifies that code is well-formed, not that it solves the problem. A kinematic check verifies that a trajectory is realisable, not that it achieves the goal. The reliable part of a grounded check is what its formalism expresses, and the formalism never expresses everything that can go wrong.
	
	The second is to ask another model. Best-of-$N$ with a learned verifier, process reward models, and LLM-as-judge frameworks all take this shape \citep{cobbe2021gsm8k,lightman2024lets,zheng2023judging}; recent work scales by ensembling model-based verifiers \citep{lifshitz2025multiagent,llmasfuser2025}. This generalises across tasks, but the verifier shares the original model's weaknesses. Errors decorrelate partially, not fully: miscalibrated confidence \citep{llmasfuser2025}, agreement that reflects shared rather than independent reasoning \citep{entanglement2026}, bias toward outputs from the same model family \citep{judgemultilingual2025}.
	
	Neither recipe is sufficient alone. \citet{kambhampati2024llmmodulo} argues that LLMs cannot reliably plan or self-verify but can be productive inside systems that integrate them with grounded checks; the position has held up across robotics \citep{ahn2022saycan,huang2022inner,liang2023codeaspolicies,song2023progprompt,grigorev2025verifyllm,spca2026}, code generation, and structured reasoning. What is missing is a principled way to combine the two when each contributes partial coverage. Recent work combines grounded and model-emitted checks in pipelines for RL training \citep{verif2025,t3rl2026}, but at test-time selection existing systems typically commit to one validator type: SayCan \citep{ahn2022saycan} multiplies an LLM utility with one learned affordance value; RoVer \citep{rover2025} applies a single learned PRM; TPVQA \citep{tpvqa2023} validates against a symbolic planner; RoboMonkey \citep{robomonkey2025} selects with one VLM. A practitioner with both a kinematic check and a VLM scorer has no off-the-shelf machinery for combining them, and the literature has not carefully ablated which pieces of such a combination actually contribute.
	
	We introduce a scoring framework for this gap. Unlike verification, which asks whether a single output satisfies a goal property (e.g.\ obstacle avoidance, reachability), scoring asks which of several candidate outputs to pick. The candidates need not include a correct one, and the framework does not certify that the chosen output is correct; it only ranks the candidates by how well they meet the available checks. For a task class, the architect fixes a backbone of universal grounded checks any acceptable output must satisfy. For a specific task, a \emph{configuring model} selects optional checks and weights for softer factors. The configuring model is kept distinct from the model producing candidates, so it cannot tilt the scoring rule toward the candidates it would itself have produced; the same conceptual gap that produces a faulty plan does not also produce a configuration that fails to look for the fault. The checks produce outputs of different types (pass/fail flags, continuous margins, judge scores), and we normalise them to a common scale before combining log-linearly, with hard vetoes (a check whose failure removes the candidate from the pool regardless of other scores) for backbone failures and soft penalties for safety-critical violations.

	We evaluate our strategy on four domains spanning binary and continuous outcomes: robot manipulation on LIBERO with SafeLIBERO augmentations, joint correctness-and-security code generation on CWEval, contamination-controlled code generation on LiveCodeBench, and retrieval-augmented QA on HotpotQA. We make four contributions:
	\begin{itemize}
		\item We build a unified test-time scoring framework that handles checks of mixed type (pass/fail, continuous, probabilistic) under one rule, with hard vetoes for the strict checks and soft penalties for the graded ones; the same scoring core runs unchanged across the four domains above despite their very different output spaces and check types.
		\item We present a component-level ablation that disentangles which parts of the pipeline improve selection and which do not.
		\item We uncover a fundamental obstacle to per-task configuration of the soft factors: a dedicated model that chooses and weights them performs no better than three simpler alternatives (uniform weights, letting the generator weight them itself, or a learned linear combiner fit to the same features), and when a single grounded factor dominates the pool, simply selecting the candidate that maximises that one factor does better still.
		\item We identify the mechanism behind this negative result: soft-factor scores cluster too tightly among the survivors of the hard checks for re-weighting to change the top pick.
	\end{itemize}

	The framework is intended as a universal scaffold that applies across the domains we consider (code generation, robot manipulation planning, retrieval-augmented QA) and is portable to others where similar checks exist. Its significance is therefore broader than any one of the four benchmarks. For robotics, it offers a principled way to let a non-negotiable physical check (reachability, collision, a keep-away margin around a fragile object or a person) override a VLM's opinion about a plan, so that a planner built from off-the-shelf components can be made safer without retraining the models involved. For software engineering, the same structure turns a collection of existing tools (compilers, test runners, static analysers, security judges) into a single selection rule, and the verdict it produces can be fed back to the generator as a structured repair signal. For retrieval-augmented systems, it gives a way to enforce that answers cite documents that exist before a judge is asked to rate them. Beyond these applications, the ablation tells practitioners building any selection pipeline where to invest effort (in eliciting reliable hard checks) and where not to (in tuning how soft factors are blended).

	\section{Related Work}
	
	\subsection{Test-time verification of LLM outputs}
	
	A substantial line of work scales test-time compute by sampling multiple candidates and selecting with a verifier. Best-of-$N$ \citep{cobbe2021gsm8k} picks the highest-scoring under a learned reward model. Verifiers have evolved from discriminative reward models to process reward models \citep{lightman2024lets,uesato2022solving}, generative verifiers \citep{zhang2024generative}, and thinking verifiers that produce verification rationales \citep{thinkprm2025,rewardingprogress2025}; surveys catalogue the design space \citep{verificationsurvey2025}. More recent work derives continuous verifier scores from scoring-token logits and scales verification by decomposing the criteria a judge is asked to check \citep{kwok2026verifier}, scores candidates jointly rather than one at a time \citep{kim2026multiseq}, and studies best-of-$N$ when only partial, per-claim verification is available \citep{zhang2026boe}. Large-scale audits of LLM judges find that high agreement and consistency coexist with severe position and format sensitivity \citep{norman2026reliability,dev2026harness}, which is one reason we confine model-emitted scores to the soft-factor pool.

	The closest related line of work is multi-verifier ensembling. Multi-Agent Verification \citep{lifshitz2025multiagent} combines aspect verifiers each prompted to check a different property; LLM-as-a-Fuser \citep{llmasfuser2025} ensembles model judgements to reduce overconfidence. Both beat a single verifier, though the verifiers they combine are all model-based and share failure modes \citep{entanglement2026}. A second wave combines grounded and model-emitted verification directly at training time: VerIF \citep{verif2025} pairs rule-based code verification with LLM verification when training instruction-following models, T${}^3$RL \citep{t3rl2026} uses code-execution evidence to upweight verified rollouts during test-time reinforcement learning updates, and Reasoning Arena \citep{zhou2026arena} applies a rule verifier first and falls back to LLM-judge tournaments only where the verifier gives no signal. AutoPyVerifier \citep{pezeshkpour2026autopy} learns a minimal set of executable checks whose conjunction approximates a target objective, and finds that the checks it retains are predominantly structural and grounded, which is consistent with our finding that the grounded factors carry the signal. Our framework adopts the same hybrid grounded-plus-model structure but treats the two sources differently: grounded checks can sit in the hard backbone and override soft factors entirely, while model-emitted checks are confined to the soft-factor pool. The test-time selection setting, that structural separation, and the ablation of which pieces contribute are what distinguish our work.

	A closely related finding from a different domain is that equal-weighted multi-dimensional rubrics underperform their best individual dimensions when dimensions differ in predictive validity \citep{criterionvalidity2026}. The mechanism (a strong dimension diluted by weak ones in a fixed-weight combiner) is structurally the same as one we report on LiveCodeBench.

	A separate line of work is the LLM-modulo framework \citep{kambhampati2024llmmodulo}: LLMs cannot reliably self-verify but can be productive inside systems with external symbolic verifiers. We extend it by treating the integration as a scoring problem rather than an iterative refinement loop, allowing checks of mixed types, and formalising how heterogeneous scores combine.
	
	\subsection{Grounded verification of LLM-generated plans in robotics}
	
	Grounding LLM outputs against the physical world has been central to LLM-based planners since their first wave. SayCan \citep{ahn2022saycan} pairs an LLM scoring skill descriptions with learned value functions for feasibility, combined multiplicatively; Grounded Decoding \citep{groundeddecoding2023} extends this to tokens; Inner Monologue \citep{huang2022inner} threads scene descriptions into the prompt; Code as Policies \citep{liang2023codeaspolicies} and ProgPrompt \citep{song2023progprompt} sidestep grounding by having the LLM write executable code.

	A second wave couples LLM planning with symbolic verification. TPVQA \citep{tpvqa2023}, ISR-LLM \citep{isrllm2024}, VerifyLLM \citep{grigorev2025verifyllm}, and SPCA \citep{spca2026} check LLM plans against classical planners, LTL representations, or compilation pipelines. In each, the validator is either a single symbolic solver or a single LLM judge.

	A third wave addresses verification for VLA models. RoVer \citep{rover2025} uses a learned PRM; SITCOM \citep{sitcom2025} a learned dynamics model; RoboMonkey \citep{robomonkey2025} a VLM verifier; SIMPACT \citep{simpact2025} a VLM coupled with a physics simulator; SelfReVision \citep{selfrevision2025} has VLMs critique their own plans. Test-time verification for VLAs has since broadened to physical-consistency gates over any policy \citep{physvla2026}, critics that select among sampled actions by pairwise comparison rather than absolute scoring \citep{vlaattc2026}, zero-shot geometric evaluation of predicted futures \citep{zhao2026geometric}, and evidence that scaling verification can outperform scaling policy learning \citep{cover2026}. Each nonetheless commits to one validator type.

	Recent VLA safety work \citep{vlasafety2026,hazardarena2026,vlsa2025,safevla2025} shows that VLAs trained on safe demonstrations can act unsafely under semantic context shift, and proposes training-time alignment, concept-based filtering, or control barrier function safety layers. Our framework is complementary: a hard backbone enforces safety conditions that learned methods cannot make non-negotiable. Work probing VLM spatial reasoning over robot motion directly \citep{mapg2026,stepeval2025,vlmmotionpref2026} shows VLMs can produce calibrated scores on specific sub-questions; we exploit this in the soft-factor pool while keeping the backbone grounded.
	
	\subsection{Decomposable utility and decision-theoretic scoring}
	
	Combining multiple scores into one ranking is classical: multi-attribute utility theory \citep{keeney1976decisions} gives the additive, multiplicative, and weighted decompositions; influence diagrams \citep{howard1984influence,shachter1986evaluating} extend Bayesian networks with decision and utility nodes; UtilityMax \citep{marom2026utilitymax} applies this to model-emitted scores. We borrow the machinery and apply it to mixed grounded and model-emitted scores, with our contribution lying in bringing heterogeneous score types to a common range and in separating configuration from generation.
	
	\section{Framework}
	\label{sec:framework}
	
	In the following, we first define the setting and the roles played by the generator, the checks, and the configuring model (\S\ref{sec:framework-setting}). We then describe the task classes, the hard-check backbone, and the pool of soft factors (\S\ref{sec:framework-tasks}), and derive the score function that combines them into a single number per candidate (\S\ref{sec:framework-score}). Finally, we describe how the score is used at inference time (\S\ref{sec:framework-inference}).

	\subsection{Components}
	\label{sec:framework-setting}
	\label{sec:framework-roles}

	Consider an LLM-based question-answering system, or a robot receiving a natural-language planning instruction. In each case, we prompt a model to produce candidate outputs, and something has to pick which one to act on. The framework divides this work among three roles, each of which is filled by a model: a \emph{drafting model} that produces the candidates, a \emph{configuring model} that decides what to check for the task at hand, and one or more \emph{scoring models} that rate properties no grounded tool can compute.

	\paragraph{Drafting model.} The drafting model (equivalently, the generator) produces candidate outputs when prompted with the task. It is the only role that touches the output space directly, and it is treated as a black box: the framework reads its candidates, and optionally its token log-probabilities for a confidence baseline, but does not modify or fine-tune it.

	\paragraph{Configuring model.} The configuring model reads the task and decides what to check. Once per task, it selects which optional factors to enable from a pre-specified pool and how to weight them, and it may supply prompt templates for the scoring models. It does not see the candidates, only the task description, so its decisions cannot be tailored to favour any particular output.

	\paragraph{Scoring models.} Scoring models rate a candidate on some property that cannot easily be computed with a grounded tool: a VLM rating semantic plausibility, or an LLM rating code quality. A scoring model takes the candidate as input, optionally with a configuration from the configuring model, and returns a score. Scoring models do not participate in training; they only score completed candidates at inference time.

	The one constraint on how the roles are assigned is that the drafting model must be distinct from the configuring model and from any scoring model used. Two failure modes motivate this. The first is direct self-bias: a model that both drafts and scores has incentive to favour its own outputs, particularly when scoring includes prompt templates the drafter would have followed. The second is shared-mode failure: even without explicit self-favouring, a model's blind spots in generation correlate with its blind spots in deciding what to check; the same conceptual gap that produces a faulty plan can produce a configuration that fails to look for the fault. The configuring and scoring roles may coincide (both sit on the evaluation side of the pipeline); only the drafter must stay separate. The constraint is at the instance level rather than the family level; the few family-level overlaps in our experiments are listed in Appendix~\ref{app:shared}.
	
	Throughout, the drafting model produces $M$ candidate outputs $\mathcal{A}_t = \{a_1, \ldots, a_M\}$ for a task instance $t$; a candidate is an answer text, a program, or a finite sequence of typed skills, depending on the domain.
	
	\subsection{Check organisation}
	\label{sec:framework-tasks}
	A central object in the framework is the \emph{task class}. A task class $T$ groups individual tasks together when they can be scored using the same set of checks: every task in the class shares which properties must hold for an output to count as acceptable, even if the specific content of each task differs. Grouping is a design decision made by the architect deploying the framework, and it matters because the backbone of checks (defined below) is fixed once per class rather than per task: every CWEval task uses the same execution and security checks even though the function to implement varies, and every LIBERO scene uses the same collision and feasibility checks even though the goal and layout vary. The four benchmarks in \S\ref{sec:experiments} are four task classes.

	For a class $T$, the architect specifies a backbone $B_T$ of checks every candidate must pass.Backbone design is manual and small-scale (three checks per domain in our experiments): the aim is to cover the non-negotiable properties, not to catalogue every failure mode. A backbone check returns a pass/fail verdict and is chosen for two properties: low false-positive rate (a passing candidate is acceptable on the property checked) and non-negotiability (violating the property invalidates the output regardless of other merits). Compilation for code, kinematic feasibility and collision absence for robot plans, and citation existence for cited factual claims are paradigm cases. A property that is safety-relevant but graded (a distance margin rather than a hard collision) belongs in the soft-factor pool instead, where it is penalised continuously rather than vetoed.

	Above the backbone sits a pool $V_T$ of \emph{soft factors}: checks that contribute graded evidence without functioning as a veto. They range from linguistic constraints (fluency), through physical margins (clearance from an obstacle, joint limits, trajectory smoothness) and task-performance signals (test pass rate, retrieval grounding), to model-emitted preferences (semantic plausibility, code quality). A soft factor is grounded (computed by a tool: a simulator, a test runner, an entailment checker) or model-emitted (a model prompted to rate a property). The pool is fixed at deployment; not every soft factor is used on every task.

	Each soft factor $v$ produces, after scaling (\S\ref{sec:framework-score}), a value $p_v(a) \in (0, 1]$ where higher is better. Constraints that are naturally continuous are mapped to this range by a monotone function of their margin. A robotics keep-away constraint, for example, is not a yes/no test: the raw signal is the minimum distance $d$ between the gripper (or carried object) and the obstacle across the planned trajectory, and the scaled factor is a monotone increasing map $p_v = \sigma\big((d - d_0)/s\big)$, where $\sigma(x) = 1/(1 + e^{-x})$ is the logistic sigmoid, near $1$ when the margin is comfortable, near $0$ as the margin closes, and centred at a task-class clearance $d_0$ with scale $s$. The same pattern covers margins to a human, file-permission strictness, or static-analysis severity: any real-valued quantity monotonically related to desirability can be scaled this way.

	For a specific task, a \emph{configuring model} (an LLM that reads the unstructured task description and emits a configuration) selects optional checks and weights for softer factors, producing a task-level specification
	\begin{equation}
		\Sigma_t = \big( V_t \subseteq V_T,\; \{w_v\}_{v \in V_t},\; \{\lambda_k\}_{k \in K_t},\; \Pi_t \big),
		\label{eq:config}
	\end{equation}
	containing the selected soft factors $V_t$, a weight $w_v > 0$ for each (how much that factor counts in the combined score), a penalty strength $\lambda_k > 0$ for each safety-critical soft constraint $k \in K_t \subseteq V_t$ (how steeply a near-violation is punished), and any prompt templates $\Pi_t$ for model-emitted scores. The configuring model is a language model rather than a small learned regressor because its input is the unstructured task description (an instruction, a docstring, a scene graph) and its job is to map that to the factors it implies: ``put the fragile bowl in the drawer'' should trigger the fragile-distance factor; a docstring mentioning a file path should trigger the path-traversal CWE pattern. The configuration is chosen once per task and reused across all $M$ candidates. The architect can override the configuring model's freedom with hard rules, for instance requiring a safety-critical factor to be enabled whenever its trigger appears in the task (a fragile-distance factor whenever the scene graph contains a fragile object), so a safety-relevant factor is never omitted on a task that needs it.
	
	\subsection{Score design}
	\label{sec:framework-score}

	The pieces so far, namely the backbone $B_T$, the soft-factor pool $V_T$, and the per-task specification $\Sigma_t$, tell us which checks to run and how much each should count. What remains is to turn the raw check outputs into a single score per candidate. Two problems: check outputs come in incompatible types (pass/fail flags, continuous margins, scaled probabilities, integer judge scores), and they have to be combined into one comparable number. We use scaling to handle the first problem and a combining function to handle the second.

	\paragraph{Score scaling.} We map each raw check output to $p_v(a) \in (0, 1]$, where higher is better, using a hand-set monotone function chosen per check type: a fixed sigmoid on the margin for binary-with-margin checks; a fixed sigmoid with a task-class reference point $d_0$ and scale $s$ for continuous margins (as in the keep-away example in \S\ref{sec:framework-tasks}); a fixed normalised log-odds map with anchor points enforced for judge scores. A cap on $w_v$ prevents any single source from dominating. This is not calibration in the statistical sense: proper calibration would require labelled data (temperature scaling \citep{guo2017calibration} for learned probabilities, or a pairwise Bradley--Terry protocol \citep{zheng2023judging} for judge outputs), which our task-class setup does not assume.

	\paragraph{Combined score.} For a candidate $a$ passing the backbone, the score is
	\begin{equation}
		S(a) = \sum_{v \in V_t \setminus K_t} w_v \log p_v(a) - \sum_{k \in K_t} \lambda_k \log c_k(a),
		\label{eq:score}
	\end{equation}
	where $c_k(a) \in (0, 1]$ measures safety satisfaction, with $c_k(a) \to 0$ indicating severe violation. The selected soft factors $V_t$, the safety-critical subset $K_t$, the weights $w_v$, and the safety-penalty strengths $\lambda_k$ are all decided per task by the configuring model and packaged into the task-level specification $\Sigma_t$ from Eq.~\ref{eq:config}. The architect fixes the allowed ranges (we use $w_v \in [0.5, 2.0]$ and $\lambda_k \in [0.5, 3.0]$ in our experiments, detailed in Appendix~\ref{app:cweval-details}); the configuring model picks specific values within those ranges. The ranges are set by hand: wide enough for the configurator to express meaningful preferences between factors (a factor-of-4$\times$ spread) and narrow enough that no single factor can numerically dominate the composite, but they are not tuned on validation data or per benchmark. The values within them are picked by the configuring model in a single zero-shot pass from the task description, with no labelled data or fine-tuning.

	The log-linear form is multiplicative aggregation of independent utilities \citep{keeney1976decisions}, the natural reading of multiple checks all of which should hold; framing candidate selection as expected-utility maximisation over a decomposed objective follows the same line as recent utility-based prompting work \citep{marom2026utilitymax}. It is unbounded below: a scaled score approaching zero on any factor drives the score arbitrarily negative, which is right when a check raises a strong objection. The same property justifies the logarithm on $c_k$: a candidate that nearly violates a safety constraint is heavily penalised relative to one that comfortably satisfies it, with the penalty growing without bound as the margin closes.
	
	\subsection{Inference-time candidate selection}
	\label{sec:framework-inference}
	
	With the roles, the backbone and soft-factor pool, the per-task specification $\Sigma_t$, and the score function in place, what remains is how they are executed when a new task instance arrives. The procedure has four steps. First, the configuring model reads the task description and produces $\Sigma_t$; this happens once per task, before any candidate is seen. Second, the drafting model emits $M$ candidates. Third, each candidate is evaluated against the backbone $B_T$, and any candidate that fails a backbone check is removed from the pool. Fourth, the survivors are scored by Eq.~\ref{eq:score} and the argmax is returned. If no candidate passes the backbone, the framework abstains; whether to resample, fall back, or escalate is the caller's decision. Candidates are scored independently, so the last two steps parallelise.
	
	\section{Experiments}
	\label{sec:experiments}

	The framework makes two claims that can only be settled empirically: that a fixed scoring core transfers across domains whose outputs and checks look nothing alike, and that each of its components earns a place in it. To test both, we want public benchmarks that differ in output space and outcome type, generators diverse enough that no single model family drives the results, and one fixed set of selection methods run unchanged on every benchmark so that ablations are comparable across domains. Table~\ref{tab:experimental-scheme} summarises the resulting scheme. This section states the design choices, reports each benchmark with one headline table and the findings that bear on the contributions, and closes with a cross-domain summary (Figure~\ref{fig:cross-domain}); settings, factor definitions, and secondary tables are in the appendices.

	\paragraph{Benchmark selection.} All four are published, community-used benchmarks: CWEval \citep{cweval2025}, LIBERO \citep{libero2024} with SafeLIBERO-style augmentations \citep{vlsa2025}, LiveCodeBench (LCB) \citep{livecodebench2024}, and HotpotQA. Each stresses the framework along a different axis: CWEval tests joint correctness-and-safety scoring; LIBERO tests transfer to a structured-plan output space with a safety-critical soft factor; LCB tests what happens when the soft-factor pool reduces to one dominant signal; HotpotQA tests whether the configurator earns its keep on continuous multi-dimensional outcomes.

	\paragraph{Generator selection.} Generators differ because domain specialisation matters: code-specialists on the code benchmarks, VLM planners on LIBERO, general instruction-tuned models on HotpotQA. Within each domain we test multiple generators (four on CWEval, LIBERO, and HotpotQA; two on LCB), fourteen conditions in total. All generators are 4-bit open-weight models that fit on a single Apple M4 with 128\,GB unified memory. The instance-level separation of \S\ref{sec:framework-roles} holds throughout; the three conditions with family-level overlap between generator and evaluator are listed in Appendix~\ref{app:shared}.

	\paragraph{Method selection.} All methods select from the same candidate pools. \textit{ours} is the full framework; \textit{top1} takes the first sample with no selection. \textit{genlogprob} picks the candidate the generator was most confident in (length-normalised log-likelihood), with no checks and no judge. \textit{backbone\_only} picks uniformly at random among candidates that pass the backbone, and \textit{judge\_only} takes the highest model-emitted judge score with no veto. Single-factor argmaxes select on one soft factor alone. Three ablations remove one component of the framework at a time: \textit{uniform} replaces the configurator's weights with uniform ones, \textit{no\_safety\_penalty} zeroes the $\lambda_k$ terms, and \textit{self\_configuring} lets the generator emit its own configuration. Finally, \textit{learned} is a global linear weighting of the same soft-factor scores fit by ridge regression with leave-one-task-out cross-validation, the strongest fixed alternative to the configurator on the same features. \textit{genlogprob} and \textit{learned} are omitted on LIBERO, where the symbolic executor exposes no log-probabilities and the combiner would require offline re-scoring. Appendix~\ref{app:shared} gives precise definitions.

	\begin{table}[h]
		\centering
		\small
		\begin{tabular}{llrl}
			\toprule
			Benchmark & Task & $N$ & Generators \\
			\midrule
			CWEval & Code + security & 25 & Qwen2.5-Coder, Qwen3-Coder, Codestral, DeepSeek-Coder \\
			LIBERO & Robot plans & 15 & Qwen2.5-VL, InternVL3, Gemma-4, Pixtral \\
			LiveCodeBench & Competitive code & 150 & Qwen2.5-Coder, Codestral \\
			HotpotQA & Multi-hop RAG & 50 & gpt-oss, Phi-4, Gemma-4, Qwen3 \\
			\bottomrule
		\end{tabular}
		\caption{Experimental scheme: benchmark, task type, number of task instances $N$, and generator families. Model sizes and quantisation are given in the appendices.}
		\label{tab:experimental-scheme}
	\end{table}

	\subsection{LLM code generation on CWEval}
	\label{sec:exp-llm}

	We begin with the domain the framework was designed for. Code generation offers the richest supply of grounded checks (a program either parses, runs, and passes a test or it does not), and the property that matters in deployment, that code be both correct and secure, is a joint outcome no single check captures. CWEval scores both axes on the same tasks, which lets us ask whether blending functional and security signals beats either alone, and whether per-task configuration of that blend is what makes the difference.

	CWEval \citep{cweval2025} provides 25 Python tasks across 19 CWE families, each with a function specification, a functionality test suite, and a CWE-specific security test suite; the joint outcome is the Effective Safety Rate, $\textnormal{ESR} = \textnormal{Functionality} \times \textnormal{Safety}$, which is zero whenever either factor is zero. Four 4-bit code-specialist generators from distinct families (Qwen2.5-Coder-32B, Qwen3-Coder-30B, Codestral-22B, DeepSeek-Coder-33B) each produce $M = 8$ candidates per task. The backbone applies three hard vetoes (parse, import, execute on one sample); the soft-factor pool combines grounded checks (test pass rates, static analysis, AST features) with three model-emitted judges, and the configuring model and judges are held fixed across all four generators. Appendix~\ref{app:cweval-details} gives the models, factor definitions, scaling, prompts, and seeds.

	\begin{table}[h]
		\centering
		\small
		\begin{tabular}{lcccccccc}
			\toprule
			& \multicolumn{3}{c}{\textit{top1}} & \textit{gen-} & \multicolumn{3}{c}{\textbf{ours}} & \textit{learned} \\
			\cmidrule(lr){2-4} \cmidrule(lr){6-8}
			Generator & Func. & Safety & ESR & \textit{logprob} & Func. & Safety & ESR & ESR \\
			\midrule
			Qwen2.5-Coder-32B & 0.933 & 0.800 & 0.762 & 0.737 & 0.920 & 0.840 & 0.775 & \textbf{0.782} \\
			Qwen3-Coder-30B & 0.720 & 0.660 & 0.630 & 0.680 & 0.790 & 0.940 & \textbf{0.750} & \textbf{0.750} \\
			Codestral-22B & 0.903 & 0.680 & 0.662 & 0.626 & 0.960 & 0.770 & 0.757 & \textbf{0.795} \\
			DeepSeek-Coder-33B & 0.743 & 0.560 & 0.529 & 0.465 & 0.843 & 0.790 & 0.639 & \textbf{0.679} \\
			\bottomrule
		\end{tabular}
		\caption{CWEval ($N = 25$, $M = 8$). Functionality, safety, and ESR for \textit{top1} and \textbf{ours}; ESR for the generator-confidence baseline (\textit{genlogprob}) and the learned combiner (\textit{learned}). ESR is averaged per task, so it is not the product of the column means. Bold marks the best ESR in each row.}
		\label{tab:cweval-crossgen}
	\end{table}

	\paragraph{Filter-then-score lifts ESR on every generator, most where the first sample is weakest.} Table~\ref{tab:cweval-crossgen} reports the headline comparison. The framework lifts ESR over \textit{top1} on all four generators, by $+0.120$ on Qwen3-Coder, $+0.110$ on DeepSeek, $+0.095$ on Codestral, and only $+0.013$ on Qwen2.5-Coder, whose first sample is already strong. Safety lifts move the same way, reaching $+0.28$ on Qwen3-Coder and shrinking to a single task on Qwen2.5-Coder. The relationship is a trend rather than a strict rank ordering, and it recurs on every other domain below. The framework also beats random selection among backbone survivors on all four generators and the \textit{judge\_only} baseline on three; the exception, DeepSeek, arises only because the framework abstains on 12\% of that generator's tasks while \textit{judge\_only} never abstains.

	\paragraph{The generator's own confidence is a weak selector.} \textit{genlogprob} underperforms the framework on every generator, by $0.04$ to $0.17$ ESR, and falls below random selection over backbone survivors on all four: the hard checks alone are worth more than the generator's entire confidence ranking. The failure is not a length bias; the model is simply most confident in fluent, plausible-looking code that fails the tests.

	\paragraph{The configuration layer is inert.} \textit{uniform} matches or beats the framework on three of the four generators (0.785 / 0.775 / 0.652 against the framework's 0.775 / 0.757 / 0.639 on Qwen2.5-Coder, Codestral, and DeepSeek) and trails it by under half a task on the fourth (0.735 against 0.750 on Qwen3-Coder); the configurator ran on every task, so this is a clean test. \textit{no\_safety\_penalty} is identical to the framework on all four. \textit{learned}, the data-fit combiner, matches or beats the framework on every generator (mean $\Delta = +0.021$ ESR, Table~\ref{tab:cweval-crossgen}), and it concentrates its weight almost entirely on the grounded pass-rate features while giving the LLM judges and CWE-pattern factors near-zero or negative weight. \textit{self\_configuring} beats the framework on three of four generators and trails on Qwen3-Coder, a counter-result to the separation requirement that we treat as indicative at $N = 25$ (the largest margin is roughly 1.5 tasks, and the Qwen3-Coder pass fell back to uniform weights on 20 of 25 tasks). The reason is the same one that recurs on every domain: once the backbone has filtered out the broken candidates, soft-factor scores cluster tightly enough that re-weighting rarely flips the top pick.

	\paragraph{Feeding the verdict back gives a further gain.} The framework's per-task verdict catalogues which checks failed and can be fed back to the generator. In a single-round revision loop on Qwen3-Coder, the generator receives that summary, produces a revision, and the revision replaces the incumbent only if it outscores it under the composite. ESR improves from $0.750$ to $0.850$ and abstention falls from 16\% to 4\%, because the failure summary rescues three of the four abstention tasks (each a missing-import fix). The composite comparison rejects every revision that scores worse, including two that would have regressed ground truth, so no kept revision is ground-truth worse than its predecessor. The reported gain is a lower bound (Appendix~\ref{app:cweval-details}).

	\subsection{VLM-generated robot manipulation plans on LIBERO and SafeLIBERO}
	\label{sec:exp-vlm}

	To test whether the scoring core transfers beyond code, we move to robot manipulation plans emitted by a vision-language model. Here a candidate is a sequence of parameterised skills, the grounded checks are kinematic and symbolic, and the outcome is whether a plan achieves its goal without endangering fragile objects. It is also the domain the safety-critical soft penalty was designed for: a plan can be entirely correct and still pass too close to something breakable.

	The benchmark has 15 scenes, 10 from LIBERO-Long \citep{libero2024} and 5 obstacle-augmented variants in the SafeLIBERO style \citep{vlsa2025}. The generator emits ordered sequences of parameterised skills, and a symbolic executor evaluates each plan against the scene's BDDL-derived success predicate; this measures plan-level correctness, not physical execution. The headline generator (Qwen2.5-VL-32B) produces $M = 12$ candidate plans per scene, and three further generators (InternVL3-38B, Gemma-4-26B, Pixtral-12B) are run on identical scaffolding, with the same configuring model and semantic-plausibility judge throughout. The backbone applies three hard vetoes (plan grammar, object-id resolution, IK feasibility); the soft-factor pool combines three grounded factors, a safety-critical fragile-distance factor, and the judge. Appendix~\ref{app:libero-details} gives the scene set, skill library, scaling, and prompts.

	\begin{table}[h]
		\centering
		\small
		\begin{tabular}{lcccccc}
			\toprule
			& \multicolumn{3}{c}{\textit{top1}} & \multicolumn{3}{c}{\textbf{ours}} \\
			\cmidrule(lr){2-4} \cmidrule(lr){5-7}
			Generator & Succ. & Safe. & Joint & Succ. & Safe. & Joint \\
			\midrule
			Qwen2.5-VL-32B & 0.93 & 0.87 & 0.80 & \textbf{1.00} & 0.87 & \textbf{0.87} \\
			InternVL3-38B & 1.00 & 0.87 & 0.87 & 0.93 & \textbf{0.93} & 0.87 \\
			Gemma-4-26B & 1.00 & 0.87 & 0.87 & 1.00 & 0.87 & 0.87 \\
			Pixtral-12B & 0.87 & 0.93 & 0.80 & 0.80 & 0.87 & 0.73 \\
			\bottomrule
		\end{tabular}
		\caption{LIBERO and SafeLIBERO (15 scenes, $M = 12$). Success, safety, and joint for \textit{top1} and \textbf{ours} across four generators. Bold marks where ours improves on \textit{top1}.}
		\label{tab:libero-crossgen}
	\end{table}

	\paragraph{The framework's value is narrow and migrates with the candidate pool.} On the headline generator, methods cluster into three tiers: random selection over survivors (0.73 joint), single-signal heuristics including \textit{top1} (0.80), and every method that uses the executor-preconditions signal (0.87). The framework's $+0.07$ over \textit{top1} traces to a single scene where \textit{top1}'s plan violates an executor precondition; safety is capped at 0.87 for every method because two scenes have no safe plan in their pool. Across the other generators (Table~\ref{tab:libero-crossgen}) the added value moves with the pool. Under InternVL3, \textit{top1} already reaches 1.00 success, so the framework's margin moves to the safety axis (0.93 against 0.87) through one scene where it prefers a candidate that uses a non-fragile object over the exact symbolic match; the \textit{no\_safety\_penalty} ablation flips that choice back, one of only two places in the study where $\lambda_k$ changes a selection (the other is a two-task gap on Phi-4 in \S\ref{sec:exp-rag}). Under Gemma~4 all methods tie because the generator emits byte-identical plans on 13 of 15 scenes. Under Pixtral the framework lands one scene below \textit{top1}, the loss falling entirely in the safety-critical subset (0.40 against 0.60 on the five SafeLIBERO scenes).

	\paragraph{A single grounded factor reproduces the framework.} \textit{uniform}, \textit{no\_safety\_penalty}, \textit{self\_configuring}, and a single-factor argmax on executor preconditions all match the full framework on every headline scene, and across all four generators the framework's joint score equals the uniform-weights joint score exactly. The executor-preconditions argmax matches the framework on three generators and beats it on Pixtral (0.87 against 0.73). On this benchmark a single well-chosen grounded factor captures the achievable signal, and the framework's separation from single-shot sampling is bounded by how diverse the candidate pool is and how strong the unfiltered baseline already is on the safety-critical slice.

	\subsection{Functional code generation on LiveCodeBench}
	\label{sec:exp-lcb}

	To check whether the CWEval findings hold for plain functional code generation, and at a sample size large enough to tighten the configurator comparison, we evaluate on LiveCodeBench \citep{livecodebench2024}, a contamination-controlled benchmark of competitive-programming problems with hidden tests. There is no security dimension, so we report functionality alone; the soft-factor pool reduces to functionality-relevant signals, which makes this the domain in which to ask what happens when one grounded factor dominates.

	We use 150 problems (35 easy, 45 medium, 70 hard) with contest dates after the training-data cutoffs of both generators; functionality is the fraction of hidden tests passed by the selected candidate, with abstention counted as zero. Two generators (Qwen2.5-Coder-32B, Codestral-22B) each produce $M = 8$ candidates per problem on the CWEval scaffolding with the security factors disabled, so the pool contains the test pass rate, two code-structure factors, and the code-quality judge. Appendix~\ref{app:lcb-details} gives the problem manifest and the differences from CWEval.

	\begin{table}[h]
		\centering
		\small
		\begin{tabular}{lcccccc}
			\toprule
			Generator & \textit{top1} & \textit{genlog-} & \textit{func\_} & \textit{uniform} & \textbf{ours} & \textit{learned} \\
			& & \textit{prob} & \textit{only} & & & \\
			\midrule
			Qwen2.5-Coder-32B & 0.481 & 0.463 & \textbf{0.676} & 0.643 & 0.656 & 0.665 \\
			Codestral-22B & 0.380 & 0.376 & \textbf{0.601} & 0.568 & 0.587 & 0.573 \\
			\bottomrule
		\end{tabular}
		\caption{LiveCodeBench ($N = 150$, $M = 8$). Functionality (fraction of hidden tests passed) per selection method; \textit{func\_only} is the argmax of test pass rate alone. Bold marks the best in each row.}
		\label{tab:lcb-headline}
	\end{table}

	\paragraph{The largest lifts in the code domain, but a single factor beats the framework.} The framework lifts functionality over \textit{top1} by $+0.175$ on Qwen2.5-Coder and $+0.207$ on Codestral (Table~\ref{tab:lcb-headline}), the largest single-condition lifts among the code benchmarks, concentrated on medium and hard problems where the first sample most often fails (Appendix~\ref{app:lcb-details}, Table~\ref{tab:lcb-difficulty}). Yet the simplest possible baseline, the argmax of test pass rate alone, beats the configured framework on both generators and at every difficulty level. On CWEval, where the pool contains both functional and security signals, the framework beats every single-factor argmax; here one factor carries essentially all the signal, and blending it with weaker ones (complexity, defensive coding, the code-quality judge) dilutes the result. Where one factor dominates, weighting it together with peripherally-correlated others hurts.

	\paragraph{The configuration null holds at $N = 150$.} Framework, \textit{uniform}, and \textit{self\_configuring} land within noise of one another on both generators, a substantially tighter test than CWEval's $N = 25$. \textit{genlogprob} is again the weakest selector, below \textit{top1} and below random selection over survivors. The learned combiner tracks the framework within $0.02$ at every difficulty level, but on this benchmark it is a near-circular check: the held-out public-test pass rate is one of its input features, so it collapses onto that feature and reproduces the single-factor argmax. It corroborates CWEval without being independent evidence.

	\subsection{Retrieval-augmented QA on HotpotQA}
	\label{sec:exp-rag}

	The three benchmarks so far have binary outcome metrics, and on all three the soft factors saturated among backbone survivors. This section asks whether continuous multi-dimensional outcomes give the configurator something to act on. It is also the domain where the safety property is semantic rather than executable: an answer must not assert claims the retrieved documents do not support.

	We use 50 questions from the HotpotQA distractor split (25 bridge, 25 comparison) with retrieval fixed in advance; the generator produces an answer with inline citations against the retrieved context, and a held-out judge distinct from every model in the pipeline scores functionality (answer equivalence to gold) and safety (no unsupported claims). Four generators (gpt-oss-20B, Phi-4, Gemma-4-26B, Qwen3-32B) each produce $M = 8$ candidates. The backbone applies three hard vetoes (answer well-formed, citations resolve, answer shape matches the question type); the soft-factor pool combines three grounded factors with three model-emitted judges (faithfulness, relevance, completeness), of which faithfulness is safety-critical. Appendix~\ref{app:rag-details} gives definitions and the ground-truth protocol.

	\begin{table}[h]
		\centering
		\small
		\begin{tabular}{lcccccccc}
			\toprule
			& \multicolumn{3}{c}{\textit{top1}} & \textit{gen-} & \multicolumn{3}{c}{\textbf{ours}} & \textit{learned} \\
			\cmidrule(lr){2-4} \cmidrule(lr){6-8}
			Generator & Func. & Safety & Joint & \textit{logprob} & Func. & Safety & Joint & Joint \\
			\midrule
			gpt-oss-20B & 0.64 & 0.88 & 0.58 & 0.55 & 0.84 & 0.93 & 0.79 & \textbf{0.83} \\
			Phi-4 & 0.80 & 0.935 & 0.735 & 0.798 & 0.76 & 0.93 & 0.713 & \textbf{0.763} \\
			Gemma-4-26B & 0.82 & 0.97 & 0.80 & 0.825 & 0.84 & 0.97 & \textbf{0.82} & 0.80 \\
			Qwen3-32B & 0.82 & 0.975 & 0.80 & 0.78 & 0.86 & 0.965 & \textbf{0.83} & \textbf{0.83} \\
			\bottomrule
		\end{tabular}
		\caption{HotpotQA distractor ($N = 50$, $M = 8$). Functionality, safety, and joint for \textit{top1} and \textbf{ours}; joint for \textit{genlogprob} and \textit{learned}. Bold marks the best joint in each row.}
		\label{tab:rag-headline}
	\end{table}

	\paragraph{The four generators span the framework's full range of value.} Table~\ref{tab:rag-headline} reports the headline comparison. On gpt-oss-20B the framework lifts joint by $+0.21$ over \textit{top1}, the largest lift in the study, driven almost entirely by functionality (0.84 against 0.64): the first sampled answer is frequently wrong or degenerate, the backbone removes it, and selection promotes a correct, grounded one. On Gemma-4-26B and Qwen3-32B the lift collapses to $+0.02$ and $+0.03$, within noise at $N = 50$, because the first sample is already strong. Phi-4 breaks the pattern: its first sample is mid-strength, yet the framework lands one task below \textit{top1}. Safety is already saturated across methods on this generator, so the lever the framework usually pulls does nothing, and its pick on the functionality axis costs two tasks. The per-question-type breakdown (Appendix~\ref{app:rag-details}, Table~\ref{tab:rag-bytype}) confirms that the lift is largest where the first sample fails most often (gpt-oss-20B comparison questions, $+0.30$) and at or below \textit{top1} wherever the first sample is already strong.

	\paragraph{The configuration null replicates, and generator confidence is unstable.} Per-task configuration does not beat uniform weights on any of the four generators (0.79 / 0.713 against 0.81 / 0.743 on gpt-oss-20B and Phi-4, exact ties on Gemma-4-26B and Qwen3-32B), the safety-penalty term is inert except for a two-task gap on Phi-4, and \textit{self\_configuring} is competitive with or slightly above the separated framework throughout. The learned combiner matches or beats the framework on three of four generators (mean $\Delta = +0.018$ joint), with its largest gain on Phi-4; unlike LCB, this comparison is not circular, since the combiner's input features are independent of the ground-truth metric. \textit{genlogprob}, meanwhile, is the single best method on Phi-4 and Gemma-4-26B but collapses on gpt-oss-20B, where the most-confident answers are short, document-grounded, and confidently wrong. Generator self-confidence is reliably the weakest selector on code and mixed on QA; it is not stable across domains.

	\subsection{Summary across domains}
	\label{sec:exp-summary}

\begin{figure}[h]
		\centering
		\includegraphics[width=\textwidth]{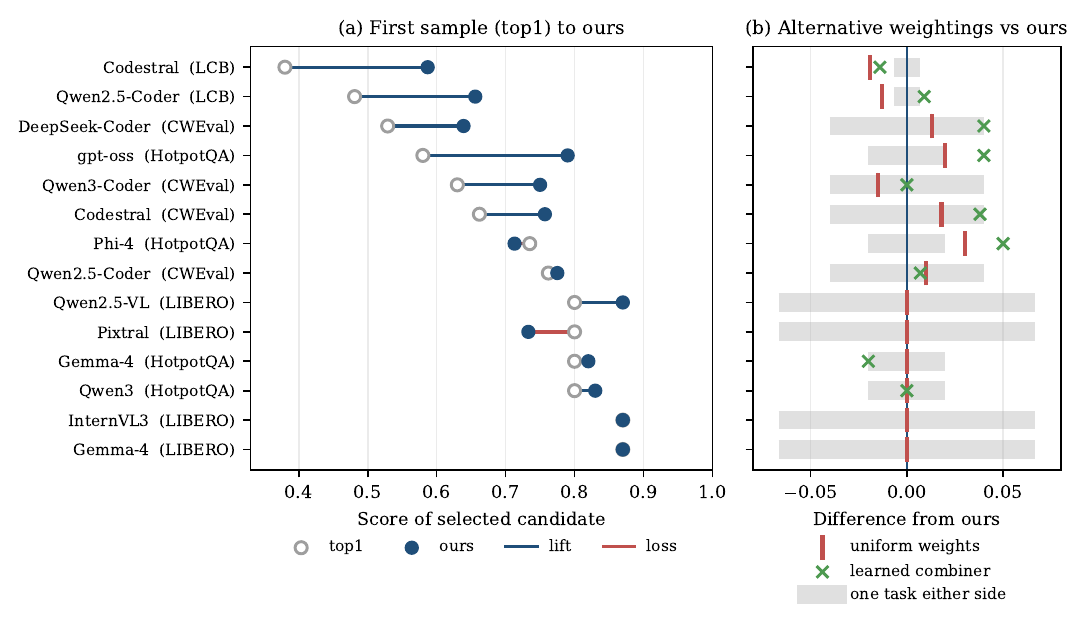}
		\caption{Cross-domain summary over fourteen generator conditions, one per row, sorted by \textit{top1} score. (a) Score of the selected candidate under \textit{top1} (open) and ours (filled); the metric is ESR on CWEval, joint on LIBERO and HotpotQA, and functionality on LCB. (b) Difference from ours for uniform weights and the learned combiner (not run on LIBERO); the grey band spans one task either side of zero.}
		\label{fig:cross-domain}
	\end{figure}

	Figure~\ref{fig:cross-domain} collects the fourteen generator conditions into two views that correspond to the paper's two main findings, one row per condition, ordered from the weakest first sample at the top to the strongest at the bottom. Panel (a) draws each condition as a segment from the \textit{top1} score to the framework's score. The segments are long ($+0.09$ to $+0.21$) in the top six rows, where the first sample scores below about 0.67, shorten steadily as the first sample approaches 0.8, and reverse in two rows (Phi-4 on HotpotQA, Pixtral on LIBERO) where the unfiltered baseline was already strong on the axis the framework acts on. The ordering mixes all four domains, as expected if the gain comes from filtering and comparison rather than anything domain-specific. Panel (b) plots, on the same rows, how far uniform weights and the learned combiner land from the framework, against a grey band marking one task either side. Most markers sit within one task of zero and all within three; the band is cleared only by the LCB rows, where one task is just $0.007$, by uniform weights on Phi-4, and by the learned combiner's small gains on gpt-oss-20B and Phi-4, with the CWEval learned gains sitting at the band's edge. How the soft factors are weighted does not change which candidate is picked.

	\section{Discussion}
	\label{sec:discussion}

	The experiments leave a consistent picture: the grounded backbone and candidate comparison carry the gain, and the per-task configuration layer does not. This section draws out what that implies for generality, for two deployment hazards the experiments exposed, for how the configuration layer could be extended, and for the study's limitations.

	\paragraph{Generality.} The scoring core was written once for CWEval and reused unchanged on LIBERO, LiveCodeBench, and HotpotQA; only the domain plug-ins (backbone checks, soft-factor pool, configurator prompt) changed. The hard/soft split held in every domain: compilation, IK feasibility, hidden-test execution, and citation resolution all fit the ``non-negotiable, cheap to check'' backbone shape, while test pass rates, collision margin, preconditions, VLM judgements, faithfulness, relevance, and completeness all fit the ``graded, informative but not decisive'' soft-factor shape. Adapting to a new domain reduces to identifying which checks belong on which side.

	\paragraph{Soft-factor misalignment is a deployment hazard.} A soft factor can look like a good proxy for the outcome and still pull selection the wrong way. Trajectory smoothness was anti-correlated with execution safety under OSC\_POSE control on an early physics-grounded LIBERO run not otherwise reported here: smoother plans gave the controller less time to settle, raising contact risk. The framework cannot detect such a mismatch on its own, so each soft factor's score has to be checked to move the right way against the outcome the deployment cares about.
	
	\paragraph{The soft-factor weighting layer has little headroom.} Two readings of the configurator null are consistent with the results across all four domains. First, the per-task variation the configurator emits is not where the value lies. Second, the weighting layer has very little headroom over uniform anyway: mean $\Delta$(learned $-$ uniform) is $+0.015$ on CWEval and $+0.005$ on RAG, both smaller than the configurator-vs-uniform spread.

	\paragraph{Automated structure discovery.} Per-task configuration is a constrained version of automated structure discovery: the configuring model selects among a pre-specified pool rather than proposing new checks. The constraint keeps the pool auditable and bounds the configuring model's failure modes; relaxing it, so that the model proposes new grounded checks and their implementations, is a natural follow-up.

	\paragraph{Limitations.} The framework requires that some grounded checks exist; for purely open-ended tasks the backbone is empty and the framework reduces to model-only scoring. The framework uses hand-set monotone maps rather than statistical calibration to bring heterogeneous scores to a common range; the combined score is only as well-founded as the reference points chosen for each check, and no better than the worst-scaled component (typically a model-emitted one). Per-task configuration adds latency that is non-trivial on short tasks with small $M$; tight-latency deployments may need pre-computed presets. Finally, the framework's separation from single-shot sampling is bounded by candidate-pool diversity, which it does not measure; on LIBERO one generator's pool collapsed to a single effective candidate. Diagnostics on within-task candidate variation are worth running before drawing conclusions about a deployment's gain.
	
	\section{Conclusion}

	We built a test-time scoring framework that brings together three functional components. A backbone of grounded hard checks plays the role of gatekeeper: it removes any candidate that fails a testable, non-negotiable property, regardless of how the candidate scores on anything else. A pool of soft factors of heterogeneous type (pass/fail, continuous margins, judge ratings) plays the role of evidence: each is scaled to a common range and combined log-linearly into one number per surviving candidate, with safety-critical factors entering as soft penalties. A configuring model, kept separate from the generator, plays the role of adapter: it reads the task and decides which soft factors to enable and how much each should count. The framework improves selection precisely where confident-but-wrong outputs are most costly: code that must be both correct and secure, robot plans that must reach a goal without endangering fragile objects, and retrieval-grounded answers that must not assert unsupported claims. Its two innovations relative to prior test-time verification are the treatment of checks of mixed type under one rule, with hard vetoes and soft penalties structurally separated, and the separation of the model that configures scoring from the model that produces candidates.

	This is demonstrated by experiments on four public benchmarks and fourteen generator conditions, in which the same scoring core ran unchanged across output spaces as different as Python programs, skill sequences, and cited answers. The framework lifts joint correctness-and-safety by up to $+0.21$ and code-generation safety by up to $+0.28$ over single-shot sampling, with a lift that grows with how weak the first sample is and that comes from the same mechanism in every domain: the backbone removes candidates the generator's own confidence would have chosen, and comparison promotes a correct, safe survivor. Feeding the framework's verdict back to the generator for one round of revision adds a further $+0.10$ ESR without ever regressing ground truth.

	Equally significant is what the component-level ablation shows does not matter. The per-task configuration layer, the component that looks most sophisticated, gives the same picks as uniform weights, as a data-fit linear combiner, and as letting the generator configure itself, and where one grounded factor dominates the pool a single-factor argmax on that factor is better still. The reason is structural: once the backbone has filtered the pool, soft-factor scores cluster too tightly among the survivors for any re-weighting to change the top pick. The practical lesson is to invest in eliciting reliable hard checks and in comparing candidates at all, and to keep the weighting layer simple. The broader lesson is methodological: test-time scoring frameworks tend to be presented as wholes, and taking ours apart showed that half of it could be removed without loss. We recommend others do the same before claiming a new mechanism earns its keep. Letting the configuration layer propose new grounded checks rather than weight a fixed pool is the natural next step, and our results suggest its value will lie in the checks it proposes rather than in how it weights them.

	\bibliographystyle{plainnat}
	\bibliography{references}

	\appendix

	\section{Shared experimental conventions}
	\label{app:shared}

	\paragraph{Selection methods.} All methods select from the same $M$ candidates per task. \textit{ours} runs the full pipeline of \S\ref{sec:framework-inference}. \textit{top1} takes the first sampled candidate with no selection. \textit{genlogprob} selects the argmax of length-normalised generator log-likelihood, with no backbone and no judge, so it can select a candidate the backbone would have removed. \textit{judge\_only} likewise ignores the backbone entirely and selects the candidate with the highest model-emitted judge score, with no veto and no grounded factor. \textit{backbone\_only} selects uniformly at random among backbone survivors, seeded by the SHA-256 of the task ID for reproducibility. Single-factor argmaxes (\textit{functionality\_only}, \textit{security\_only}, \textit{collision\_only}, \textit{precondition\_only}, \textit{faithfulness\_only}, \textit{relevance\_only}) select the backbone survivor with the highest value of one named soft factor. The three ablations take the framework and remove one component: \textit{uniform} (written \textit{no\_configuring\_model} in the run logs) uses a fixed configuration with $V_t = V_T$, $w_v = 1$, $\lambda_k = 1$; \textit{no\_safety\_penalty} takes the emitted $\Sigma_t$ but overrides every $\lambda_k$ to 0; \textit{self\_configuring} has the drafting model emit $\Sigma_t$ on a separate prompt context in place of the configuring model. Every method that uses the backbone abstains when no candidate passes it; abstention is scored as functionality 0 and, where a safety axis exists, safety 1.

	\paragraph{Learned combiner.} \textit{learned} is a single global linear weighting of the scaled soft-factor scores, fit by ridge regression with leave-one-task-out cross-validation on the per-candidate ground-truth outcome, operating in the same log-linear space and over the same backbone survivors as the framework. Only the weights differ from the framework: learned from data rather than emitted per task by the configurator. On HotpotQA the per-candidate ground truth is recomputed from the held-out \texttt{gemma-3-27b} judges. The combiner was not run on LIBERO, where it would require offline re-scoring of every candidate.

	\paragraph{Generator--evaluator overlap.} The instance-level separation of \S\ref{sec:framework-roles} holds in every condition, but three conditions have family-level overlap between generator and an evaluation-side model, which we note for transparency. On LIBERO, \texttt{Qwen2.5-VL-32B} generates while \texttt{Qwen2.5-VL-7B} judges semantic plausibility; this is the closest pairing. On CWEval, the Qwen-family generators are scored in part by a \texttt{Qwen3-8B} security judge. On HotpotQA, \texttt{Gemma-4-26B} generates while \texttt{gemma-3-12b} configures and \texttt{gemma-3-27b} scores ground truth.

	\section{CWEval experimental details}
	\label{app:cweval-details}

	\paragraph{Models.} Four generators from distinct families, all 4-bit code-specialist models: \texttt{Qwen2.5-Coder-32B-Instruct}, \texttt{Qwen3-Coder-30B-A3B-Instruct}, \texttt{Codestral-22B-v0.1}, and \texttt{DeepSeek-Coder-33B-Instruct}. Held constant across all four: configuring model \texttt{gpt-oss-120b-MXFP4-Q4}; code-quality judge \texttt{gemma-3-12b}; security judge \texttt{Qwen3-8B-4bit}; adjudicator \texttt{gpt-oss-120b}.

	\paragraph{Backbone checks.} A candidate is removed from consideration if any of the following fails. \textit{Parse}: the candidate parses as valid Python. \textit{Import}: the candidate's module imports without raising; the check captures import errors and any exception raised during top-level execution. \textit{Execute}: the candidate's entry point is called once with sample arguments from the task's test fixtures; the check passes if the call returns without raising.

	\paragraph{Soft factors: grounded.} Each grounded factor produces a raw value mapped to $p_v(a) \in (0, 1]$.

	\textit{Test pass rate.} Fraction of CWEval's functionality test suite the candidate passes; identity map to $[0, 1]$ with a small floor of $10^{-3}$ to avoid $\log 0$ in Eq.~\ref{eq:score}.

	\textit{Security pass rate.} Fraction of CWEval's CWE-specific security test suite the candidate passes; identity map with floor as above. Safety-critical.

	\textit{Static analysis.} \texttt{bandit} is run on the candidate; the raw score is the count of findings weighted by severity (low: 1, medium: 3, high: 9). Scaling: $p_v = \exp(-\text{weighted\_count} / \tau)$ with $\tau = 5$. Safety-critical.

	\textit{CWE-family regex hits.} A small table of regex patterns per CWE family (e.g.\ for CWE-022, patterns matching naive path concatenation without normalisation). The raw score is the count of pattern matches; scaling is $\exp(-\text{count} / 2)$. Safety-critical. The regex table covers 11 of the 25 task CWEs and returns no-penalty for the rest.

	\textit{AST complexity.} Combined LOC plus AST node count plus cyclomatic branch count, normalised against a per-task-class reference, sigmoid-mapped with steepness 1.0. Lower complexity scores higher.

	\textit{Defensive coding patterns.} Count of input validation, exception handling, and sanitisation patterns detected via AST traversal (try-except blocks, isinstance checks, explicit type coercion, string escaping). Scaling: $\sigma(\text{count} / 3)$.

	\paragraph{Soft factors: model-emitted.} Three judges run as separate prompts, each producing a 0--10 integer against anchored descriptions at 0, 5, and 10. The code-quality judge (\texttt{gemma-3-text-12b-it-4bit}) rates how cleanly the candidate addresses the task; the security judge (\texttt{Qwen3-8B-4bit}, safety-critical) rates whether the targeted vulnerability is mitigated; the adjudicator (\texttt{gpt-oss-120b-MXFP4-Q4}, safety-critical) applies stricter joint functionality-and-security criteria with explicit rejection conditions. Scaling for all three is a normalised log-odds map from the integer score to $(0, 1]$ with anchor points enforced.

	\paragraph{Configuring model interface.} The configurator (\texttt{gpt-oss-120b-MXFP4-Q4}) receives the task's function signature, docstring, target CWE ID, and a list of available soft factors with their score-type metadata. It produces a JSON object containing: the enabled subset $V_t$ of soft factors, weights $w_v$ in $[0.5, 2.0]$, safety-penalty strengths $\lambda_k$ in $[0.5, 3.0]$ for safety-critical factors, and a free-text rationale. The configurator runs once per task; the resulting $\Sigma_t$ scores all $M$ candidates for that task.

	\paragraph{Revision loop.} The revision loop of \S\ref{sec:exp-llm} replays the candidate sets and verdicts behind Table~\ref{tab:cweval-crossgen}. Six of 25 tasks keep the revision: the three abstention rescues plus three composite wins that leave ground truth unchanged. Functionality rises from $0.790$ to $0.910$; safety falls from $0.940$ to $0.920$ because abstention is scored as safe by convention. The remaining abstention (\texttt{cwe\_732\_2}) is an infrastructure artefact: the revision is ground-truth correct, but our execute check runs in a sandbox that denies a file-permission operation the CWEval harness permits. The source run persists each task's enabled factors and all candidate composite scores, but not the configurator's weights. We therefore score the revision against the full set of weightings consistent with the stored scores, a polytope pinned by the framework and no-safety-penalty verdicts, and keep it only when it outscores the incumbent under every such weighting. Ambiguity resolves to the incumbent, so the reported gain is a lower bound: it forfeits at most $+0.02$ ESR, on one task whose rejected revision was ground-truth better.

	\paragraph{Sampling.} $M=8$ candidates per task at temperature 0.7, top-$p$ 0.95, 2048 token budget on the generator, with deterministic per-candidate seeds derived from the global seed 0. Configurator and judge budgets are 4096 and 1024 tokens respectively.

	\section{LIBERO and SafeLIBERO experimental details}
	\label{app:libero-details}

	\paragraph{Scene set.} Ten scenes from LIBERO-Long (the \texttt{libero\_10} suite) and five SafeLIBERO-style obstacle-augmented scenes. Scene-graph extraction (object poses, bounding boxes, names) runs against the LIBERO MuJoCo simulator; execution then routes through a symbolic executor that evaluates each plan against a hand-translated success predicate in the symbolic-state vocabulary (\texttt{in\_container}, \texttt{on\_surface}, \texttt{is\_open}). Per-scene metadata declares this predicate and the fragile- and obstacle-marked objects. Three of the ten LIBERO-Long predicates approximate their BDDL originals (two drop a ``turn on stove'' sub-goal absent from the symbolic vocabulary, one drops a spatial-relation clause); the approximations are documented in the manifest.

	\paragraph{Skill library.} Six parameterised skills: \texttt{pick}, \texttt{place}, \texttt{push}, \texttt{pour}, \texttt{open}, \texttt{close}. Each declares required waypoints for IK feasibility, preconditions over the scene graph, and a symbolic effect that updates the scene graph for downstream preconditions. Under the symbolic executor, each skill applies its effect when its preconditions are satisfied in the current symbolic state; mis-ordered or precondition-violating steps cause the executor to abort the step and the resulting state to fail the success predicate.

	\paragraph{Backbone checks.} \textit{Plan grammar}: the candidate parses as JSON conforming to the skill schema. \textit{Object existence}: every referenced \texttt{object\_id} resolves to a scene-graph entry, including virtual fixture entries. \textit{IK feasibility}: each skill's required waypoints have a Franka Panda IK solution within joint limits, queried via \texttt{robosuite}'s IK wrapper.

	\paragraph{Soft factors.} \textit{Collision-margin integral.} For each waypoint in each skill, the signed distance to the nearest obstacle-marked object is computed. Inter-skill straight-line transitions between consecutive skills are sampled at 10 intermediate points each. The plan's collision margin is the minimum across all samples; scaling is a centred sigmoid $\sigma(s(m - m_0))$ with $m_0 = 0.05$~m and $s = 30$.

	\textit{Executor pre-conditions.} The plan is symbolically executed against the scene graph; the score is the fraction of skills whose preconditions are satisfied when reached. Identity map to $[0, 1]$.

	\textit{Trajectory smoothness.} Sum of inter-segment direction-change angles across the IK-solved trajectory; the score is $\exp(-\text{total\_angle} / \tau)$ with $\tau$ set per task class.

	\textit{Fragile-distance margin.} Same as collision-margin but over fragile-marked objects, with $m_0 = 0.10$~m, $s = 30$. Safety-critical, enters Eq.~\ref{eq:score} as a $c_k$ term.

	\textit{Semantic-plausibility judge} (\texttt{Qwen2.5-VL-7B-Instruct-4bit}). The judge receives the instruction, the scene-graph summary as text, and the candidate plan as JSON, and produces a 0--10 score against anchors (0: does not attempt the instruction; 5: attempts it with clear issues; 10: accomplishes it cleanly).

	\paragraph{Models, configuration, and sampling.} Generators: \texttt{Qwen2.5-VL-32B-Instruct-4bit} (headline), \texttt{InternVL3-38B-4bit}, \texttt{gemma-4-26b-a4b-it-4bit}, and \texttt{Pixtral-12B}. The configurator (\texttt{gemma-3-12b}, text-only) receives the instruction and a structured scene-graph summary and emits $\Sigma_t$ with the same fields as on CWEval; its prompt enforces that the fragile-distance factor is enabled whenever a fragile-marked object appears in the scene graph. Single-factor argmaxes target collision margin and executor pre-conditions. $M=12$ candidates per scene at temperature 1.0, top-$p$ 0.95, 1024 token budget, seeded as on CWEval.

	\paragraph{Success and safety scoring.} Success is the symbolic success predicate evaluated against the symbolic-state output of the symbolic executor, for both LIBERO-Long and SafeLIBERO scenes. Safety is the indicator that no skill in the selected plan would bring the gripper or carried object within an unsafe margin of any fragile-marked or obstacle-marked object, evaluated geometrically against the IK-solved waypoints. Abstention is counted as success $= 0$, safety $= 1$.

	\section{LiveCodeBench experimental details}
	\label{app:lcb-details}

	The setup reuses the CWEval scaffolding (Appendix~\ref{app:cweval-details}): same backbone, same configurator and code-quality judge, same sampling settings. Only the differences are stated here.

	\paragraph{Problem manifest.} 150 problems drawn from LiveCodeBench release v6, filtered to contest dates 2024-08-03 through 2025-04-06 to lie after the training-data cutoffs of both code generators in this section. Difficulty mix as released: 35 \texttt{easy}, 45 \texttt{medium}, 70 \texttt{hard}. Platforms: \texttt{atcoder} (stdin/stdout I/O) and \texttt{leetcode} (functional return-value tests). Each problem ships with a hidden test suite (public plus private tests, capped at 12 per problem for evaluation cost); functionality is the fraction of hidden tests passed by the selected candidate, executed against the same host harness used in CWEval.

	\paragraph{Soft factor pool.} The CWEval pool restricted to the functionality-relevant subset: \texttt{lcb\_test\_passrate} (the analogue of CWEval's test pass rate, computed against a held-out portion of the public tests), \texttt{code\_complexity}, \texttt{defensive\_coding}, and the code-quality judge (\texttt{gemma-3-12b}). The security factors and the adjudicator from CWEval are registered but disabled via the configuration; \textit{security\_only} is therefore not a valid method and is omitted. \textit{judge\_only} uses the code-quality judge alone, and \textit{functionality\_only} is the single-factor argmax on \texttt{lcb\_test\_passrate}.

	\paragraph{Results by difficulty.} Table~\ref{tab:lcb-difficulty} breaks the headline result down by problem difficulty, as means across the two generators. On easy problems all methods cluster because \textit{top1} already reaches 0.83; the framework's lift is largest on medium ($+0.27$) and narrows on hard ($+0.19$), where even good selection can only recover candidates that exist in the pool. Per generator, framework / uniform / self-configuring is 0.656 / 0.643 / 0.666 on Qwen2.5-Coder and 0.587 / 0.568 / 0.597 on Codestral. The learned combiner's weight on \texttt{lcb\_test\_passrate} is $+0.32$ and $+0.28$ on the two generators, with the other three features near zero.

	\begin{table}[h]
		\centering
		\small
		\begin{tabular}{lccc}
			\toprule
			Method & easy ($n=35$) & medium ($n=45$) & hard ($n=70$) \\
			\midrule
			\textit{top1} & 0.830 & 0.444 & 0.222 \\
			\textit{genlogprob} & 0.786 & 0.478 & 0.199 \\
			\textbf{ours} & 0.919 & 0.711 & 0.415 \\
			\textit{learned} & 0.912 & 0.695 & 0.423 \\
			\textit{functionality\_only} & \textbf{0.921} & \textbf{0.723} & \textbf{0.443} \\
			\bottomrule
		\end{tabular}
		\caption{LiveCodeBench functionality by difficulty, means across the two generators. Bold marks the best per column.}
		\label{tab:lcb-difficulty}
	\end{table}

	\section{HotpotQA experimental details}
	\label{app:rag-details}

	\paragraph{Benchmark.} 50 questions sampled from the HotpotQA distractor validation split, balanced across question types (25 bridge, 25 comparison). For each question, the supporting paragraphs plus distractor paragraphs from the original distractor split are passed as the retrieval context; retrieval is fixed at manifest preparation. Keyfacts for completeness scoring are extracted from gold answers via a one-shot \texttt{Qwen3-8B} prompt at manifest preparation, producing per-question bulleted fact lists.

	\paragraph{Backbone.} \textit{Answer non-degenerate}: the answer is non-empty and is not a refusal phrase from a closed list (``I don't know'', ``cannot answer'', etc.). \textit{Citations resolve}: every cited document ID in the answer exists in the per-question retrieval set. \textit{Answer-question compatible}: a permissive regex check that yes/no questions receive yes/no answers; entity questions receive entity-shaped answers; permissive default on ambiguous cases.

	\paragraph{Soft factors.} \textit{Citation density}: number of distinct documents cited divided by answer length in tokens, sigmoid-mapped (centre 1.5/100 tokens). \textit{Retrieval coverage}: intersection of cited documents with the supporting-fact set from HotpotQA gold, divided by the supporting-set size; identity on $[0, 1]$. \textit{Answer length normalised}: bell curve on word count centred at the per-question-type expected length. \textit{Faithfulness} (safety-critical): the judge scores each sentence in the answer against its cited document(s) as SUPPORTED or UNSUPPORTED; the factor is the supported fraction. \textit{Answer relevance}: ternary judge YES/PARTIAL/NO mapped to $1.0$/$0.5$/$\epsilon$. \textit{Completeness}: per-keyfact COVERED/NOT\_COVERED ratio from the keyfact extraction.

	\paragraph{Models and configuration.} Generators: \texttt{gpt-oss-20B}, \texttt{Phi-4}, \texttt{Gemma-4-26B}, and \texttt{Qwen3-32B}. The configurator and the faithfulness and relevance judges are \texttt{gemma-3-12b}; the completeness judge is \texttt{Qwen3-8B}. The configurator receives the question, its type (bridge or comparison), and the soft-factor metadata, and its prompt enforces that faithfulness is enabled with $\lambda_k > 0$ on every task. Single-factor argmaxes are \textit{faithfulness\_only} and \textit{relevance\_only}.

	\paragraph{Ground-truth scoring.} A held-out judge (\texttt{gemma-3-27b}, distinct from every model in the pipeline) scores functionality (answer equivalence to gold) and safety (no claims unsupported by any document in the retrieval set, cited or not). The safety metric is deliberately stricter than the framework's faithfulness factor, which checks only cited claims against cited documents.

	\paragraph{Additional results.} Table~\ref{tab:rag-bytype} breaks the headline joint score down by question type. The learned combiner's margin over the framework is $+0.04$, $+0.05$, $0.00$, and $-0.02$ on gpt-oss-20B, Phi-4, Qwen3-32B, and Gemma-4-26B respectively. Qwen3-32B is the only condition where \textit{backbone\_only} (0.77) falls below \textit{top1} (0.80): the pool is uniformly strong, so the backbone removes little and a random survivor can be worse than the first sample. On the same generator \textit{judge\_only} and \textit{self\_configuring} edge the framework (0.840 against 0.830), because the backbone checks citation resolution rather than hallucination directly and a pure judge score can avoid a marginally more hallucinatory survivor. \textit{genlogprob} on gpt-oss-20B falls to 0.48 bridge functionality: its most-confident answers are short, document-grounded, and wrong.

	\begin{table}[h]
		\centering
		\small
		\begin{tabular}{llccc}
			\toprule
			Generator & Split & \textit{top1} joint & \textbf{ours} joint & $\Delta$ \\
			\midrule
			\multirow{2}{*}{gpt-oss-20B}
			& Bridge & 0.640 & 0.760 & $+0.120$ \\
			& Comparison & 0.520 & 0.820 & $+0.300$ \\
			\midrule
			\multirow{2}{*}{Phi-4}
			& Bridge & 0.820 & 0.780 & $-0.040$ \\
			& Comparison & 0.651 & 0.647 & $-0.004$ \\
			\midrule
			\multirow{2}{*}{Gemma-4-26B}
			& Bridge & 0.860 & 0.900 & $+0.040$ \\
			& Comparison & 0.740 & 0.740 & $0.000$ \\
			\midrule
			\multirow{2}{*}{Qwen3-32B}
			& Bridge & 0.840 & 0.920 & $+0.080$ \\
			& Comparison & 0.760 & 0.740 & $-0.020$ \\
			\bottomrule
		\end{tabular}
		\caption{HotpotQA joint score by question type (25 bridge, 25 comparison) for \textit{top1} and \textbf{ours}, with the difference $\Delta$.}
		\label{tab:rag-bytype}
	\end{table}

\end{document}